\documentclass[11pt]{article}
\PassOptionsToPackage{numbers}{natbib}
\usepackage[square,sort]{natbib}
\usepackage[left=1in,right=1in,top=1in,bottom=1in]{geometry}

\usepackage{microtype}
\usepackage[colorlinks=true,citecolor=blue,urlcolor=blue]{hyperref}
\usepackage{url}
\usepackage{booktabs}

\usepackage{inconsolata}
\usepackage[table,xcdraw,dvipsnames]{xcolor}
\usepackage{xspace}
\usepackage{graphicx}
\graphicspath{../images/}
\usepackage[skip=0pt]{subcaption}
\usepackage{wrapfig}
\usepackage[framemethod=tikz]{mdframed}
\usepackage{multicol}
\usepackage{multirow}

\usepackage[noend]{algorithmic}
\usepackage{algorithm}

\makeatletter
\newcommand\footnoteref[1]{\protected@xdef\@thefnmark{\ref{#1}}\@footnotemark}
\makeatother

\usepackage{amsmath}
\usepackage{amssymb}
\usepackage{mathtools}
\usepackage{amsthm}
\theoremstyle{plain}

\theoremstyle{definition}

\usepackage[inline]{enumitem}
\setenumerate[1]{itemsep=0pt,partopsep=0pt,parsep=\parskip,topsep=5pt}
\setitemize[1]{itemsep=0pt,partopsep=0pt,parsep=\parskip,topsep=5pt}
\setdescription{itemsep=0pt,partopsep=0pt,parsep=\parskip,topsep=5pt}

\usepackage[capitalise]{cleveref}
\Crefname{equation}{Eq.}{Eqs.}
\Crefname{figure}{Fig.}{Figs.}
\Crefname{table}{Tab.}{Tabs.}
\Crefname{algorithm}{Alg.}{Algs.}
\Crefname{section}{Sec.}{Secs.}
\Crefname{appendix}{App.}{Apps.}
\Crefname{theorem}{Thm.}{Thms.}
\Crefname{remark}{Rmk.}{Rmks.}

\newcommand\tmv[2]{ ${#1} {\scriptstyle \pm {#2}}$}

\definecolor{lightblue}{rgb}{0.68, 0.85, 0.9}
\definecolor{lightgreen}{rgb}{0.82, 1.0, 0.74}
\definecolor{lightred}{rgb}{1.0, 0.8, 0.8}
\definecolor{lightgray}{rgb}{0.85, 0.86, 0.84}

\newif\ifacl

\title{
Synthesizing and Adapting Error Correction Data for Mobile Large Language Model Applications
}

\author{
Yanxiang Zhang\thanks{Equal contribution. Reverse alphabetical order.},\;\;Zheng Xu$^*$,\;\;Shanshan Wu$^*$,\;\;Yuanbo Zhang,\;\;Daniel Ramage \vspace{0.5em}\\
\textit{Google}\vspace{0.5em}\\
\texttt{\{zhangyx, xuzheng, shanshanw, zyb, dramage\}@google.com}
}

\date{}
\begin{document}
\setboolean{acl}{false} 

\maketitle

\begin{abstract}
Error correction is an important capability when applying large language models (LLMs) to facilitate user typing on mobile devices. In this paper, we use LLMs to synthesize a high-quality dataset of error correction pairs to evaluate and improve LLMs for mobile applications. 
We first prompt LLMs with error correction domain knowledge to build a scalable and reliable addition to the existing data synthesis pipeline. 
We then adapt the synthetic data distribution to match the mobile application domain by reweighting the samples. 
The reweighting model is learnt by predicting (a handful of) live A/B test metrics when deploying LLMs in production, given the LLM performance on offline evaluation data and scores from a small privacy-preserving on-device language model.
Finally, we present best practices for mixing our synthetic data with other data sources to improve model performance on error correction in both offline evaluation and production live A/B testing. 
\end{abstract}

\section{Introduction}

\begin{figure*}[thb]
\centering
\includegraphics[width=0.9\linewidth]{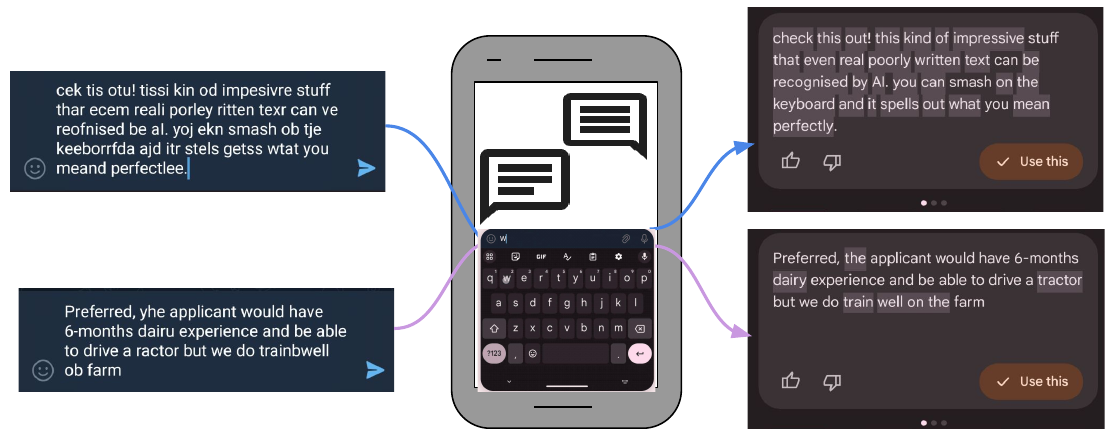}
\caption{\small Examples of mobile LLM applications for error correction. User typing data has a domain shift compared to public web data. LLMs rewrite and correct highly corrupted text based on the context of the input itself.} \label{fig:teaser}
\end{figure*}


Modern typing applications on mobile devices use many machine learning models, e.g., language models (LMs)\ifthenelse{\boolean{acl}}{~\citep{ouyang2017mobile,liu2024proofread}}{\citep{ouyang2017mobile,hard2018federated,xu-etal-2023-federated,liu2024proofread}}. The generative capacity of LMs can significantly improve user experience by (automatically) correcting various errors and predicting next words to facilitate typing. Recent advancement in large language models (LLMs) have achieved impressive performance on many language tasks~\citep{openai2024gpt4, geminiteam2024, llama2024}, opening new opportunities for rewriting in mobile applications\ifthenelse{\boolean{acl}}{~\citep{gunter2024apple,liu2024proofread}}{~\citep{gunter2024apple,liu2024proofread,zhu2023towards}}. In practice, LLMs can be deployed on mobile devices or on servers in datacenters. However, mobile devices have limited resources that currently only support moderate-sized LLMs (often less than 10 billion parameters). Even for LLMs on servers, moderate-sized models are preferred for mobile applications because of the considerations of latency, privacy and serving cost. 

Error correction (EC) is an important capacity of LLMs for mobile applications (see examples in \cref{fig:teaser}). As LLMs' general capacity can decrease with the model size~\citep{wei2022emergent,cho2024heterogeneous}, it is important to evaluate and improve moderate-sized models for mobile applications. 
Moreover, the data distribution of mobile applications can differ from commonly collected public web data~\citep{hard2018federated,xu-etal-2023-federated,wu2024prompt}; typing on mobile touchscreens introduce more errors~\citep{shi2025simulating} in addition to common grammatical errors~\citep{bryant2023grammatical,stahlberg2021synthetic}. Such EC data for mobile applications differs from much of current LLMs' training data. 

Post-training with high-quality data is commonly used to align LLMs with users~\citep{wei2021finetuned,chung2022scaling,ouyang2022training} and bridge the domain shift~\citep{cho2024heterogeneous}. Low-Rank Adaptation (LoRA) method, which only trains a small subset of parameters, is efficient for fine-tuning models for mobile applications~\citep{hu2022lora}. LoRA additionally provides the flexibility to fine-tune a set of different adapters to customize for various downstream tasks, useful for deploying LLMs on mobile devices\ifthenelse{\boolean{acl}}{~\citep{gunter2024apple}}{~\citep{cho2024heterogeneous,gunter2024apple}}. However, collecting high-quality data for post-training for mobile applications is challenging because of the domain shift and privacy considerations on user data. 

Production LLM mobile applications have developed pipelines to synthesize error correction data. \citet{liu2024proofread} collects public web data, and then uses trained task-specific models\ifthenelse{\boolean{acl}}{~\citep{lichtarge2020data}}{~\citep{lichtarge2020data,stahlberg2021synthetic}} to detect grammatical errors. A typing simulator adds more mobile-specific errors to construct EC pairs based on the web data with detected grammar errors. These EC data pairs are split into training and validation datasets. This data pipeline extracts only a small set of EC data from a large collection of web data due to the detection and selection process. Moreover, the data distribution of web data differs from the mobile user distribution, as discussed in~\citep{wu2024prompt}. Indeed, we observe a discrepancy between offline evaluation on validation data and live A/B test metrics in production.

Privacy-preserving methods are required to access in-domain user data to improve the model performance. Federated learning (FL), where devices collaboratively learn a model without transferring user data, and differential privacy (DP), where model is mathematically guaranteed not to memorize training data, are combined to privately fine-tune LMs~\citep{xu-etal-2023-federated,choquette2024amplified,mcmahan2024hassle}. However, production DP FL systems on mobile devices only reliably train models with 10 million parameters~\citep{daly2024federated}. Differentially private synthetic data is another promising approach to collect high-quality privacy-preserving data\ifthenelse{\boolean{acl}}{~\citep{kurakin2023harnessing,yue2022synthetic}}{~\citep{yu2024privacy,kurakin2023harnessing,tan2025synthesizing,xie2024differentially,hou2024pre,yue2022synthetic}}. However, DP synthetic data generation requires iterative interaction between LLMs and private data, such as fine-tuning LLMs as data generators. The quality of synthetic data also decreases with the generator model size.
These methods are not yet applied to training moderate-sized LLMs with billions of parameters for production mobile applications.

In this paper, we synthesize error correction data to improve LLMs with billions of parameters for mobile applications. 
In a production data pipeline, we incorporate human knowledge of the mobile application domain and grammar errors to carefully design prompts, and use LLMs instead of grammar error detectors to scalably and reliably synthesize EC pairs (\cref{sec:prompt}).
To further overcome the discrepancy between offline evaluation on (synthetic) EC data and live A/B test metrics for model deployment in practice, we propose to adapt the data distribution to match the mobile application domain by reweighting the samples (\cref{sec:adapt}). Small LMs with less than 100 million parameters are fine-tuned by federated learning with differential privacy on user data. These small LMs are used to generate initial scores for each offline evaluation sample. A reweighting model is parameterized to predict a final score for each sample based on the initial small LM scores. As the number of intial LM scores is small, the lightweight reweighting model is learnt by reweighting per-sample evaluation to predict only a handful of A/B test metrics collected during model deployment. We demonstrate that the reweighting model, together with privacy-preserving small LMs, effectively predicts live A/B test metrics. 
Finally, we present best practices for mixing our synthetic data with other data sources to improve the model performance (\cref{sec:mix}). LoRA method is used to further fine-tune an LLM with billions of parameters that is already post-trained for general purpose instruction following. A continue training strategy, where the model is first fine-tuned on our large-scale synthetic data, followed by fine-tuning on a mixture of existing smaller dataset and reweighted synthetic data, achieves superior performance on various offline evaluations, and 2.47\% to 7.18\% relative improvements on key metrics in production live A/B test. 

\ifthenelse{\boolean{acl}}{
\section{Synthesizing Error Correction Data} 
}{
\section{Synthesizing Error Correction Data with LLMs} 
}\label{sec:prompt}

\ifthenelse{\boolean{acl}}{}{\begin{table*}[ht]
\small
    \centering
\begin{tabular}{@{}p{\textwidth}@{}}
\toprule
{\cellcolor[HTML]{EFEFEF} \textbf{LLM prompt template}} \\ 
Imagine that you are an English school teacher. Your goal is to teach high school student English grammar. Here are some common grammartical errors: \colorbox{lightgray}{[...]} \par
\\
Given the following sentence(s): \par
\colorbox{lightred}{[...original sentence...]} \par
What grammartical errors are the students likely to make? \par
\\
Now apply these grammatical errors to the original sentence(s), and generate the ungrammatical sentence(s). Do not modify the original sentence(s) except applying the grammatical errors. \par
\\
The output should be in the following format: \colorbox{lightgray}{[...]} \par
\\
Finally, correct the grammatical errors in the generated ungrammatical sentence(s). Do not modify the sentence(s) except correcting the grammatical errors. The output should be in the following format: \colorbox{lightgray}{[...]}
\\ \midrule
{\cellcolor[HTML]{EFEFEF} \textbf{Example}} \\
\colorbox{lightred}{Original sentence}: Yesterday I went to a store that has nice furniture. \par
\colorbox{lightblue}{LLM output}:\par
**Ungrammatical sentences**: Yesterday I went to a store that have nice furnitures. \par
**Error 1: Subject-verb agreement error**: ``have'' should be ``has'' to agree with the singular subject ``store''.\par
**Error 2: Plurization error**: ``furnitures'' should be ``furniture'' as it is an uncountable noun.\par
**Corrected sentences**: Yesterday I went to a store that has nice furniture.
\\ \bottomrule
\end{tabular}
    \caption{Our LLM prompt for adding grammar errors and an example output. The prompt starts with a role (i.e., school teacher), analyzes the original sentence, and adds potential grammar errors, followed by correcting all errors. The \colorbox{lightgray}{[...]} parts are omitted for simplicity.}
    \label{tab:llm_prompt}
\end{table*}}

\begin{figure}[bth]
\centering
\begin{subfigure}[b]{0.45\linewidth}
\centering
\includegraphics[width=\textwidth]{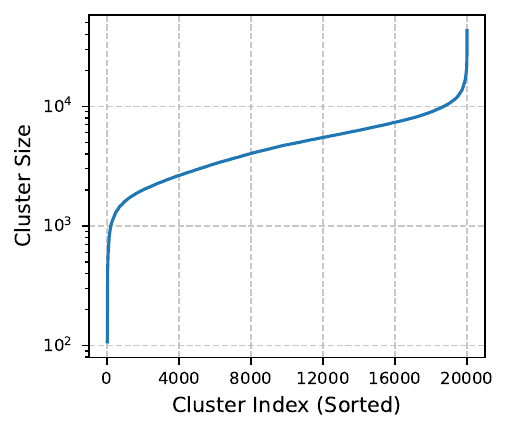}
\caption{\footnotesize }
\label{fig:cluster_sort}
\end{subfigure}
\begin{subfigure}[b]{0.45\linewidth}
\centering
\includegraphics[width=\textwidth]{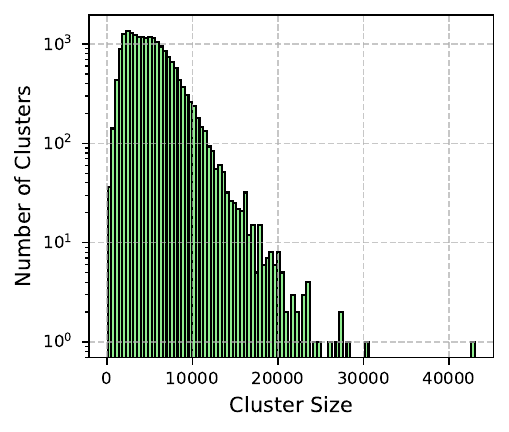}
\caption{\footnotesize }
\label{fig:cluster_hist}
\end{subfigure}
\caption{\small The statistics of the 20k clusters for 100 million documents. The mean with standard deviation of cluster sizes is $5225 \pm 2972$. } \label{fig:cluster}
\end{figure}

In this section, we discuss prompting LLMs as an addition to error correction data pipeline for efficiency and effectiveness, and show its advantage in scalability and domain adaptation.  
Following~\citet{wu2024prompt}, we synthesize an initial dataset in the domain of typing text on mobiles, by filtering and transforming public web data (i.e., C4~\citep{raffel2020exploring} dataset), and collecting LLM generations with carefully crafted prompts of human knowledge. The initial dataset contains more than 100 million documents of conversation-like text, and even a small subset (about 0.2\%) is much larger than the original EC dataset in production. We subsample the initial typing text dataset to reduce the subsequent processing costs from prompting LLM to add grammar and typing errors. To ensure good diversity and coverage during sampling, we first embed the documents using the Gecko~\citep{lee2024gecko} text embedding model, run k-means clustering to obtain 20k clusters. See \cref{fig:cluster} for statistics of clustering. We then sample 10 data points per cluster, resulting in a dataset contains about 200k documents, which has 2M examples where each example is either a sentence, or a user's utterance. Each example is relatively short similar to the examples in our target distribution, i.e., texts typed by users using their mobile keyboards in chatting or search applications. Majority of these texts are clean (i.e., error-free). 

To synthesize the EC text pairs, we add two types of errors to the clean texts: grammar error, and typing error. The grammar error is added by Gemini Ultra model~\citep{geminiteam2024}, and Table~\ref{tab:llm_prompt} shows the template of our prompt and an example. We experiment with different model sizes and find that Gemini Ultra performs best for analyzing and adding grammar errors. For high-quality data generation, the LLM is prompted to perform two more tasks in addition to generating the ungrammatical texts: 
\begin{enumerate*}[label=\color{purple}(\arabic*)]
    \item The first task is to describe the added grammar errors. This allows us to perform a global analysis of the added grammar errors, and confirm our data cover all the grammar error types from~\citep{bryant2017automatic}. The top 4 error categories (and its percentage in our synthetic data) are related to verb (52\%), missing words (15\%), plural (10\%), and capitalization (5\%). In terms of the number of grammar error per example, 12\% examples have 1 error, while more than 80\% examples have 2 or 3 errors.
    \item The second task is to correct the ungrammatical texts with LLM added grammar errors. We only keep examples when the corrected text and the original clean text are equal. 
    This filtration process removes around 40\% of the data.
\end{enumerate*}
After adding the grammar errors, we next add typing errors that simulate the behavior of real users typing with mobile keyboard. This is done by heuristic rules that add various typing errors, such as transposition, omission, repetition, and spatial errors~\citep{liu2024proofread}.

\begin{figure}[thb]
\centering
\begin{subfigure}[b]{0.45\linewidth}
\centering
\includegraphics[width=\textwidth]{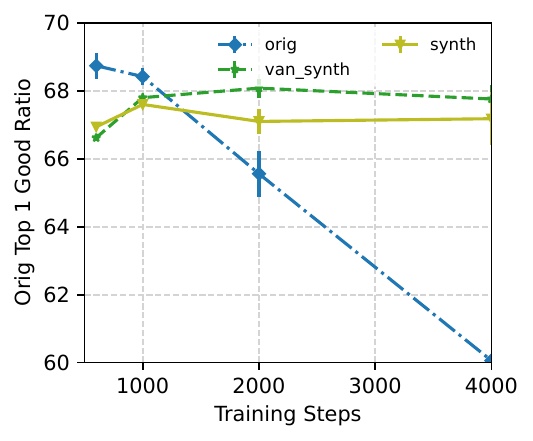}
\caption{\footnotesize }
\label{fig:top1_orig}
\end{subfigure}
\begin{subfigure}[b]{0.45\linewidth}
\centering
\includegraphics[width=\textwidth]{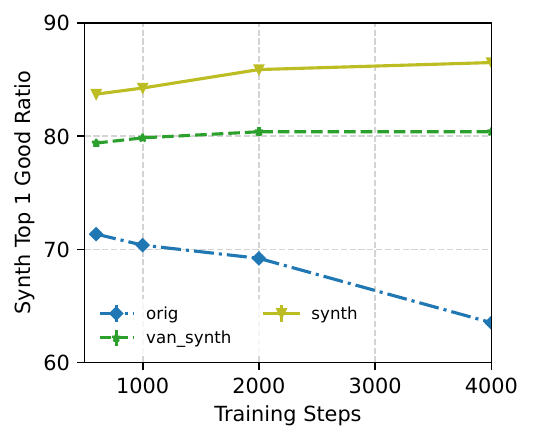}
\caption{\footnotesize }
\label{fig:top1_synth}
\end{subfigure}
\caption{\small Good ratio for error correction on the (a) original validation data  and (b) synthetic validation data. The models are trained with the original training data, synthetic training data, and vanilla sampling of synthetic data without clustering. LLMs are used to judge whether the EC output is acceptable to compute good ratio. Our large-scale LLM assisted synthetic data works well on both domains even if there is potential distribution shift from the original dataset collected by error detection on public web data.} \label{fig:top1}
\end{figure}

\subsection{Evaluation Setup and Preliminary Results} \label{subsec:eval}

Our synthetic EC dataset has about 1.2M examples. Each example is a pair of (corrupted, clean) sentences. We random sample a small subset of our synthetic data for validation, and use the rest of data for LoRA fine-tuning a Gemini Nano model~\citep{geminiteam2024}. Both the training and validation dataset are much larger than the original dataset synthesized by the previous production data pipeline. \cref{fig:seq_acc,fig:top1,fig:top3} shows the results of training and evaluation with the small original production dataset, and our large synthetic dataset, respectively. We provide an additional ablation curve on our synthetic data with vanilla subsampling instead of clustering-based subsampling. \cref{fig:seq_acc} (in \cref{app:prompt_results}) measures the error correction performance by sequence accuracy, i.e., the exact match between corrected sentence and the target clean sentences, and shows that fine-tuning help while too many steps on small dataset may quickly degrade utility. Our evaluation on error correction matches previous observation on dialogue generation and summarization tasks~\citep{cho2024heterogeneous}. 

We further use Gemini Pro models as judges to measure whether the corrected sentence is a high-quality rewrite of the target sentence even if they do not exact match for each word, and report the good ratio for the top 1 output and the best of top 3 outputs from our fine-tuned LLMs, in \cref{fig:top1,fig:top3}. Good ratio mimics the user behavior on selecting rewritten text from mobile applications. We select a small number of models from training steps $\{600, 1000, 2000, 4000\}$ for evaluation to reduce the cost of LLM judges. Models from two different training runs are evaluated to compute standard deviation for error bars. The trend of the sequence accuracy and good ratios align well \cref{fig:seq_acc,fig:top1,fig:top3}. We observe performance discrepancy between the original production dataset and our synthetic dataset, which suggests a potential domain difference. Fine-tuning on our large-scale synthetic data is more robust compared to the small original dataset, and achieves competitive model performance even when evaluated on the original validation set. The clustering-based subsampling achieves comparable results on the original evaluation, and better results on the synthetic evaluation, compared to vanilla subsampling. In the rest of the paper, we will use the synthetic data subsampled with the clusters. 

\section{Privacy-Preserving Domain Adaptation by Reweighting} \label{sec:adapt}
We have synthesized a large-scale error correction dataset in \cref{sec:prompt} by carefully prompting LLMs to simulate typing text and systematically add errors. However, \citet{wu2024prompt} suggests public LLMs and human prior knowledge in prompt may not be sufficient to bridge the potential domain shift. When deploying previously trained models, we observe misalignment in offline evaluation on the original validation set, and live A/B test metrics. We also observe the discrepancy between original validation set and our synthetic validation set in \cref{fig:top1_orig}. As our synthetic data explicitly guided LLMs with prior knowledge on mobile typing for synthesis, is it closer to the domain of mobile applications in practice?
In this section, we developed a privacy-preserving approach for domain adaptation by reweighting samples in the dataset. The reweighting model is built upon a small LM trained with DP FL, and a handful of live A/B test metrics tracked in previous model deployment.  

When evaluating an error correction model $M$ on a dataset $\{(x_i, y_i)\}_{i=1}^{N}$ of $N$ (corrupted, clean) samples, a measurement $\chi(M(x_i), y_i) \in \{0, 1\}$ is generated for each sample by comparing the model output $M(x_i)$ and corresponding target $y_i$. We have offline metric for evaluating the model by taking the average over all samples, i.e., $\sum_{i=1}^N \, \chi(M(x_i), y_i) / N$, which becomes sequence accuracy in \cref{fig:seq_acc} when $\chi(\cdot, \cdot)$ is exact match, and good ratio in \cref{fig:top1,fig:top3} when $\chi(\cdot, \cdot)$ is judged by LLMs.
To reweight samples for domain adaptation, we first train two small LMs $S_p, S_f$ of about 8 million parameters for scoring samples. Model $S_p$ is trained on public C4 dataset, and model $S_f$ is further fine-tuned from $S_p$ on user data in a production FL system~\citep{xu-etal-2023-federated,wu2024prompt}. Model $S_f$ is a privacy-preserving model with formal DP guarantee $\epsilon < 10$, and captures the domain information from mobile application. We define a reweighting model parameterized by $\theta = (\theta_f, \theta_p, \theta_b)$ as 
\ifthenelse{\boolean{acl}}{
\begin{equation} \label{eq:reweight}
\begin{split}
w(\theta, y_i) = & C_{\min} + (C_{\max} - C_{\min} \\
& ) \, \sigma(\theta_f S_f(y_i) + \theta_p S_p(y_i) + \theta_b),
\end{split}
\end{equation}
}{
\begin{equation}
w(\theta, y_i) = C_{\min} + (C_{\max} - C_{\min}) \, \sigma(\theta_f S_f(y_i) + \theta_p S_p(y_i) + \theta_b), \label{eq:reweight}
\end{equation}
}
where $C_{\min}, C_{\max}$ are constants determining the minimum and maximum value of the reweighting scores, $\sigma(\cdot)$ is the sigmoid function, and $S_f(\cdot), S_p(\cdot)$ represent the average log likelihood on predicting words in the target sentence $y_i$.

When deploying $K$ models $\{M_j(\cdot)\}_{j=1}^{K}$ in practice, we collect corresponding live A/B test metrics $\{v_j\}_{j=1}^{K}$. We consider key metrics like click through rate and accept rate for error correction in mobile applications, and hence each $v_j \in \mathbb{R}^d$ is a vector representing multiple metrics. We optimize the objective below to learn the reweighting model,

\ifthenelse{\boolean{acl}}{
{\small
\begin{align} \label{eq:opt} 
  & \min_{\theta, \alpha} R(\theta, \alpha) + \lambda \| \frac{1}{N}\sum_{i=1}^N w(\theta, y_i) -1 \|^2, \, \\ 
  & R(\theta, \alpha) = \sum_{j=1}^K\| \frac{\alpha_1}{N}\sum_{i=1}^N \, w(\theta, y_i)  \chi(M(x_i), y_i)  + \alpha_0 - v_j \|^2 \nonumber
\end{align}
}
}{
\begin{equation} \label{eq:opt}
    \min_{\theta, \alpha} R(\theta, \alpha) + \lambda \| \frac{1}{N}\sum_{i=1}^N w(\theta, y_i) -1 \|^2, \, R(\theta, \alpha) = \sum_{j=1}^K\| \frac{\alpha_1}{N}\sum_{i=1}^N \, w(\theta, y_i)  \chi(M(x_i), y_i)  + \alpha_0 - v_j \|^2
\end{equation}
}
where $\alpha=(\alpha_1, \alpha_0)$ is regression parameter to predict live metrics from offline evaluation; per-sample reweighting score $w(\theta, y_i)$ is defined in \cref{eq:reweight} to adapt offline data to mobile application domain to achieve small regression residual $R(\theta, \alpha)$; $\lambda$ is a hyperparameter on the regularizer of the reweighting scores. 

\begin{table*}[thb]
    \centering
    \small
    \begin{tabular}{c|c|c|c}
         \hline
         $R(\theta, \alpha) $ & $w=1$ & $w(y_i) \in \{0, 1\}$~\citep{wu2024prompt} & Our $w(\theta, y_i)$\\
         \hline
         Train & $1.51 \times 10^{-4}$ & $1.41 \times 10^{-4}$ &  $1.19 \times 10^{-4}$\\
         CrossVal & $(5.26 \pm 2.76) \times 10^{-5} $ &  ($4.23 \pm 2.42) \times 10^{-5} $ &  ($3.99 \pm 3.15) \times 10^{-5} $ \\
         Val & $(3.51 \pm 4.15) \times 10^{-6}$ & $(5.44 \pm 7.10) \times 10^{-5}$ &  $(2.07 \pm 2.56) \times 10^{-6}$ \\
         \hline
    \end{tabular}
    \caption{\small Regression residual $R$ for different reweighting strategies. Smaller residual suggests reweighting helps predicting live metrics. We report mean and standard deviation for predicting each live metrics $v_j$ in held-one out cross validation. We also use held-one out for validation set to fit only regression parameters with fixed reweighting.}
    \label{tab:res}
    \vspace{-0.5cm}
\end{table*}

\begin{figure}[bth]
\centering
\begin{subfigure}[b]{0.45\linewidth}
\centering
\includegraphics[width=\textwidth]{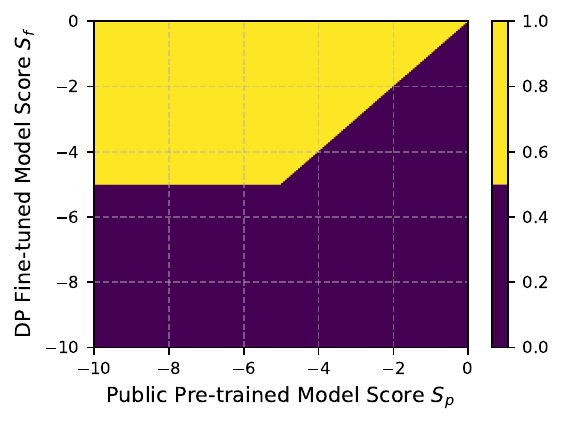}
\caption{\footnotesize }
\label{fig:colm_heat}
\end{subfigure}
\begin{subfigure}[b]{0.45\linewidth}
\centering
\includegraphics[width=\textwidth]{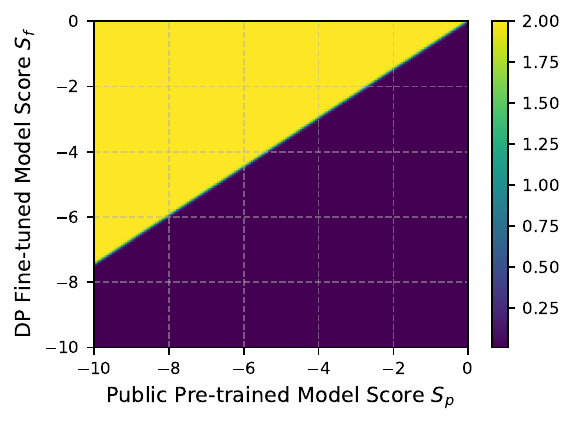}
\caption{\footnotesize }
\label{fig:reweight_heat}
\end{subfigure}
\caption{\small Comparing the (a) heuristic $\{0, 1\}$ reweighting in~\citep{wu2024prompt} and (b) our reweighting model $w(\theta, \cdot)=0.01 + 1.99 \sigma(40.64 S_f -30.44 S_p -1.59)$. Both methods use public pre-trained small LM $S_p$ and the same model further fine-tuned with DP FL $S_f$. The learnt scores in (b) have large overlap with manual selection in (a) from ~\citep{wu2024prompt}.} \label{fig:colm_reweight}
\end{figure}

We use auto differentiation and L-BFGS optimizer in JAX~\citep{jax2018github} to optimize \cref{eq:opt} to learn regression parameters $\alpha \in \mathbb{R}^{2d}$ and reweighting parameters $\theta \in \mathbb{R}^{3}$. The dimensionality of $\theta,\alpha$ is relatively small, and they can be learnt from  a handful of live metrics $\{v_j\}_{j=1}^{K}$ collected during launching different error correction models. We collected two sets of live metrics, the training set evaluated ~10 models with live A/B test, and the validation set evaluated ~5 models. We use top-3 good ratio as in \cref{fig:top3} for offline evaluation on the original dataset. Due to production test configuration, the two sets of live metrics have different scales and hence we cannot use the same regression parameter. We set the range of reweighting scores as $C_{\max}=2, C_{\min}=0.01$, and regularizer strength $\lambda=0.01$. \cref{tab:res} summarizes residuals for training, cross-validation and validation, and reweighting achieves smaller residual when predicting live metrics across different settings. The absolute value of cross-validation and validation residuals are smaller than training ressiduals as training is the summation over all live metric samples in \cref{eq:opt}.

After training, our reweighting model parameters are $(\theta_f, \theta_p, \theta_b)=(40.64, -30.44, -1.59)$, which suggests the reweighting score is positively correlated with the fine-tuned model output $S_f(\cdot)$ calibrated by pre-trained model output $S_p(\cdot)$. The difference of the two model outputs represent the likelihood discrepancy, which has also been used for inference time domain adaption~\citep{liu2024tuning} and training data detection~\citep{kandpal2023user}. \citet{wu2024prompt} discusses a heuristic filtering strategy for domain adaptation that is effective for selecting data to train small LMs for mobile applications. The heuristic filtering is equivalent to setting $w(y_i)=1$ when $S_f(y_i) > S_p(y_i)$ and $S_f(y_i) > -5$, and $w(y_i)=0$ otherwise. This heuristic approach often helps predicting live A/B test metrics compared to uniform weighting, but fails sometimes, and generally achieves higher residual than our reweighting score. \cref{fig:colm_reweight} shows the difference between our reweighting model and ~\citep{wu2024prompt} for different pre-trained and fine-tuned model scores. 

\begin{figure}[bth]
\centering
\begin{subfigure}[b]{0.45\linewidth}
\centering
\includegraphics[width=\textwidth]{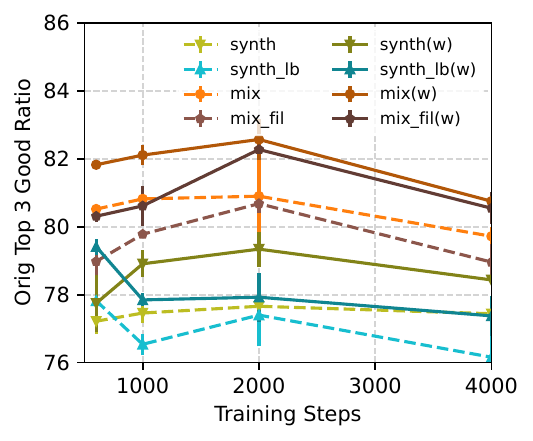}
\caption{\footnotesize }
\label{fig:top3_orig_lb}
\end{subfigure}
\begin{subfigure}[b]{0.45\linewidth}
\centering
\includegraphics[width=\textwidth]{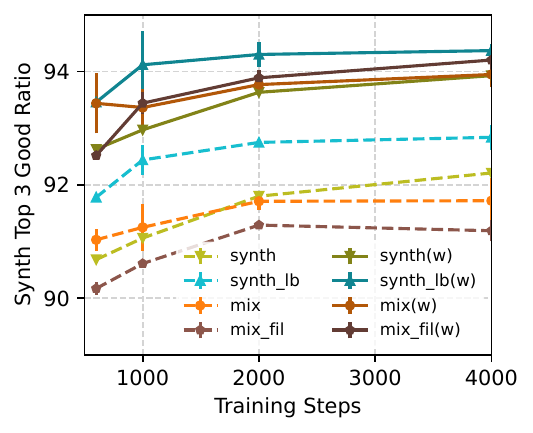}
\caption{\footnotesize }
\label{fig:top3_synth_lb}
\end{subfigure}
\caption{\small Good ratio for the best of top 3 candidates for error correction on the (a) original validation data and (b) synthetic validation data. Solid lines reweight the samples by the $w(\theta, \cdot)$ model learnt to fit live A/B test metrics in \cref{sec:adapt}. The models are trained with synthetic training data with the same setting as in \cref{fig:top3}; $\times 4$ increased batch size (synth\_lb); mixture of original and synthetic data; and mixture of original and filtered by $w(\theta, \cdot)$ (mix\_fil). LLMs are used to judge whether the error correction output is acceptable to compute good ratio. } \label{fig:top3_lb}
\end{figure}

Finally, we further apply a defense in depth strategy when using our reweighting models for privacy-preserving domain adaptation. We use a standard production PII detection pipeline to remove any possible sensitive information in the synthetic data, even if they are hallucinated by LLMs. And our fine-tuned LLMs are equipped with another layer of safety and privacy safeguarding when deploying in practice.

\section{Mixing Data for Fine-tuning} \label{sec:mix}

Based on our synthetic dataset in \cref{sec:prompt} and domain-adaptive reweighting model in \cref{sec:adapt}, we improve Gemini Nano~\citep{geminiteam2024} for error correction by LoRA fine-tuning~\citep{hu2022lora}. By combining the original dataset and the large-scale (reweighted) synthetic dataset in a continue training strategy, the model performance is improved in both offline evaluation  and live A/B test in production. Unless otherwise specified, our experiments use the same configuration for training previous production models with the original data, as described in \cref{sec:prompt}. As Gemini Nano is already post-trained with general purpose instructions, our fine-tuning is a continuous post-training. During inference after model deployment, our model is only effective for mobile applications when our fine-tuned LoRA module is applied to the base Gemini Nano model.

\begin{table*}[thb]
    \centering
    \footnotesize
    \setlength{\tabcolsep}{4pt}
    \begin{tabular}{c|c|c|c|c|c|c|c|c}
         \hline
         Training & \multicolumn{4}{c|}{Original Data Eval (\%)} & \multicolumn{4}{c}{Synthetic Data Eval (\%)} \\
         \cline{2-9}
         Method & Top-1 & Top-1 (w) & Top-3 & Top-3 (w)  & Top-1 & Top-1 (w) & Top-3 & Top-3 (w) \\
         \hline
            Original\ifthenelse{\boolean{acl}}{}{~\citep{liu2024proofread}} & \tmv{68.74}{0.38} & \tmv{71.16}{0.54} & \tmv{79.34}{0.02} & \tmv{80.38}{0.26} &
                \tmv{71.35}{0.31} & \tmv{76.24}{0.11} & \tmv{82.96}{0.30} &  \tmv{86.24}{0.26}\\
            SynthLB & \tmv{66.64}{1.28} & \tmv{68.25}{1.26} &\tmv{77.4}{0.91} & \tmv{77.93}{0.72} &
                \tmv{\underline{87.5}}{0.12} & \tmv{\underline{90.29}}{0.07} & \tmv{\underline{92.75}}{0.05} & \tmv{\underline{94.30}}{0.21} \\
            Mix & \tmv{\underline{70.22}}{0.94} & \tmv{\underline{72.91}}{0.31} & \tmv{\underline{80.9}}{1.22} & \tmv{\underline{82.57}}{0.61} &
             \tmv{85.37}{0.13} & \tmv{88.71}{0.30} & \tmv{91.71}{0.15} &  \tmv{93.77}{0.07} \\
         \hline
            ContOrig & \tmv{68.82}{0.14} & \tmv{71.21}{0.27} & \tmv{79.34}{0.54} & \tmv{80.66}{0.13}  & \tmv{77.40}{0.34} & \tmv{80.78}{0.15} & \tmv{86.49}{0.57} & \tmv{89.35}{0.64} \\
            ContMix & \tmv{69.22}{0.26} & \tmv{71.52}{0.26} & \tmv{79.86}{0.26} & \tmv{81.33}{0.19}  & \tmv{\textbf{86.04}}{0.32} & \tmv{88.88}{0.06} & \tmv{\textbf{92.03}}{0.29} & \tmv{93.69}{0.19} \\
            ContMixFil & \tmv{\textbf{70.48}}{0.00} & \tmv{\textbf{73.31}}{0.10} & \tmv{\textbf{80.78}}{0.46} & \tmv{\textbf{82.28}}{0.87}  & \tmv{85.78}{0.20} & \tmv{\textbf{89.51}}{0.02} & \tmv{91.71}{0.25} & \tmv{\textbf{93.90}}{0.14} \\
         \hline
    \end{tabular}
    \caption{\small Good ratio for error correction on the original validation data and synthetic validation data. Top-3 evaluates the best of three model outputs. Columns with ``(w)'' reweight the samples by the $w(\theta, \cdot)$ model learnt to fit live A/B test metrics in \cref{sec:adapt}. The models are trained by the original dataset as in \cref{fig:top1}; the synthetic data with large batches, mixture of original and synthetic data as in \cref{fig:top3_lb}; and three continue training strategies. Continue training gets the best, or close to best performance on offline evaluation of both original and synthetic validation data.  }
    \label{tab:cont}
    \vspace{-0.2cm}
\end{table*}

We discuss our model training practices and observations. 
\begin{enumerate*}[label=\color{purple}(\arabic*)]
\item \textbf{Increasing batch size.} Our synthetic dataset is much larger than the original dataset. As shown in \cref{fig:seq_acc,fig:top1,fig:top3}, at 4000 steps, the model performance trained on synthetic data still increases on synthetic validation data, while only starts to saturate on original validation data. In fact, 4000 steps do not complete a single epoch on our synthetic training data. We increase the batch size to $\times 4$, which is the largest batch size without requesting more resources. We found our LoRA fine-tuning is relatively robust for learning rate between $\times 1$ and $\times 4$ of the original learning rate, and hence fixed the learning rate to be $\times 1$. As shown in \cref{fig:seq_acc_lb,fig:top1_lb,fig:top3_lb}, large batch training achieves comparable performance on original validation data, while improves the performance on synthetic validation data. We use large batch training in following experiments.
\item \textbf{Simple mixing } the small original dataset and the large scale synthetic data improves the performance on original validation data, but slightly degrades the performance on synthetic validation data. There is a trade-off on the ratio of the mixture: while it is relatively robust when we have original and synthetic ratio in the range of $1:1$ and $1:8$, ratio $1:4$ achieves good balance and is used in the following mixing experiments. 
\item \textbf{Reweighting for training and evaluation.} Reweighting model in \cref{sec:adapt} is used to adapt the offline data distribution to the mobile application distribution. The trend in reweighted metrics in \cref{fig:top1_lb,fig:top3_lb} generally align with the uniform weighted counterparty. In addition to reweighting to bring offline evaluation closer to live A/B test metrics, we further explore reweighting for domain adaptation in training. We filter the synthetic data set and only keep samples with reweighting scores $w(\theta, y_i) \geq w_{t}$. We choose the threshold $w_t=1$ as $w(\theta, y_i)$ is in the range of $C_{\min}=0.01$ and $C_{\max}=2$, and about half of the samples in the synthetic dataset have reweighting scores passed the threshold. We only filter our synthetic dataset as the original dataset is already very small. Mixing the filtered synthetic data with the original dataset for training achieves good reweighted metrics even if the uniform weighted metrics slightly degrades compared to mixing with the full synthetic data. 
\item \textbf{Continue training.} As our synthetic data is large, we propose a continue training strategy: first fine-tune on the full synthetic dataset for $1000$ steps (about one epoch), and then continue training on the original data (ContOrig), the mixture of original and synthetic dataset (ContMix), and the mixture of the original and filtered synthetic data (ContMixFil), see \cref{tab:cont}. For each training method, we select the best model from steps $\{600, 1000, 2000, 4000\}$, and run training at least twice to compute the standard deviation.
\end{enumerate*}

In \cref{tab:cont}, ContMix and ContMixFil achieve best or close to best results on both original validation data and synthetic validation data. They achieve higher good ratio on the original validation data than model trained on the original data only, or the mixture of original and synthetic data. They are comparable to the best performance on synthetic validation data achieved by training on synthetic data only with large batches. 
As ContMixFil achieves better performance on the reweighted metrics that better reflects the mobile application domain, we further compare the model trained by Original and ContMixFil in production live A/B test.
Compared to Original, ContMixFil achieves 2.47\% to 7.18\% relative improvement on key production metrics like click through rate and accept rate.

\ifthenelse{\boolean{acl}}{\section{Conclusion}}{\section{Conclusion and Discussion}}
This paper presents a method to enhance error correction in mobile LLMs by creating a high-quality synthetic dataset using LLM prompts enriched with domain knowledge. We further adapt the public (synthetic) data to better match the domain of production mobile applications by developing a privacy-preserving reweighting model, using a small LM trained with federated learning and differential privacy, alongside a few live A/B test metrics. Our experiments show that fine-tuning a billion-size LLM with a mixture of the original dataset and the reweighted synthetic data, especially via continue training, significantly improves performance in offline evaluations and live A/B tests.

\ifthenelse{\boolean{acl}}{}{Our preliminary exploration on reweighting suggests combining live data from production applications and LLMs in a privacy-preserving manner is promising, and there are a lot of possibilities with the limited accessible information. Our usage of reweighting scores in training by filtering samples considers the trade-off of effectiveness, easy-to-implement, and future maintenance in production. There are many other potential domain adaptation methods for future experiments. Finally, We leverage a small privacy-preserving LM to capture domain information from mobile applications, while other forms of information such as histogram~\citep{xie2024differentially,hou2024pre,yu2024privacy,tan2025synthesizing} are worth considering, especially given the flexibility of the next generation FL systems in trusted execution environments~\citep{daly2024federated}. With more data generated from the interaction of LLMs and users in production deployment, our approach can become more powerful for not only domain adaptation but also other improvement such as personalization and agency, enabled by privacy-preserving methods.  }

\section*{Acknowledgements}
The authors thank Felix Stahlberg, Shankar Kumar, and Michael Xuelin Huang for discussions in the early stage of the project; Zachary Garrett and Shumin Zhai for reviewing an early draft.

\bibliography{ref.bib}
\bibliographystyle{plainnat}

\newpage

\appendix

\ifthenelse{\boolean{acl}}{
\section{Prompt in \cref{sec:prompt}}
\begin{table*}[ht]
\small
    \centering
\begin{tabular}{@{}p{\textwidth}@{}}
\toprule
{\cellcolor[HTML]{EFEFEF} \textbf{LLM prompt template}} \\ 
Imagine that you are an English school teacher. Your goal is to teach high school student English grammar. Here are some common grammartical errors: \colorbox{lightgray}{[...]} \par
\\
Given the following sentence(s): \par
\colorbox{lightred}{[...original sentence...]} \par
What grammartical errors are the students likely to make? \par
\\
Now apply these grammatical errors to the original sentence(s), and generate the ungrammatical sentence(s). Do not modify the original sentence(s) except applying the grammatical errors. \par
\\
The output should be in the following format: \colorbox{lightgray}{[...]} \par
\\
Finally, correct the grammatical errors in the generated ungrammatical sentence(s). Do not modify the sentence(s) except correcting the grammatical errors. The output should be in the following format: \colorbox{lightgray}{[...]}
\\ \midrule
{\cellcolor[HTML]{EFEFEF} \textbf{Example}} \\
\colorbox{lightred}{Original sentence}: Yesterday I went to a store that has nice furniture. \par
\colorbox{lightblue}{LLM output}:\par
**Ungrammatical sentences**: Yesterday I went to a store that have nice furnitures. \par
**Error 1: Subject-verb agreement error**: ``have'' should be ``has'' to agree with the singular subject ``store''.\par
**Error 2: Plurization error**: ``furnitures'' should be ``furniture'' as it is an uncountable noun.\par
**Corrected sentences**: Yesterday I went to a store that has nice furniture.
\\ \bottomrule
\end{tabular}
    \caption{Our LLM prompt for adding grammar errors and an example output. The prompt starts with a role (i.e., school teacher), analyzes the original sentence, and adds potential grammar errors, followed by correcting all errors. The \colorbox{lightgray}{[...]} parts are omitted for simplicity.}
    \label{tab:llm_prompt}
\end{table*}
\clearpage
}{}

\section{Additional results in \cref{sec:prompt}} \label{app:prompt_results}

\begin{figure}[htb]
\centering
\begin{subfigure}[b]{0.45\linewidth}
\centering
\includegraphics[width=\textwidth]{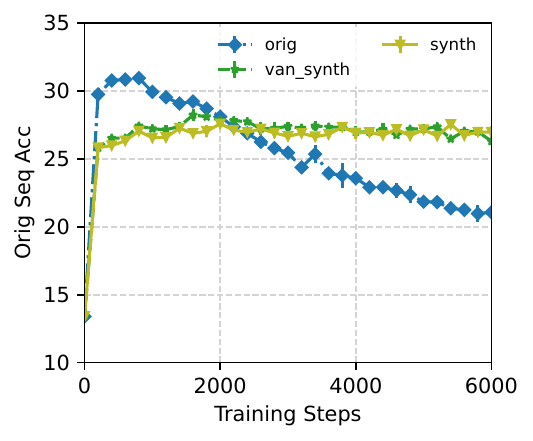}
\caption{}
\label{fig:seq_acc_orig}
\end{subfigure}
\begin{subfigure}[b]{0.45\linewidth}
\centering
\includegraphics[width=\textwidth]{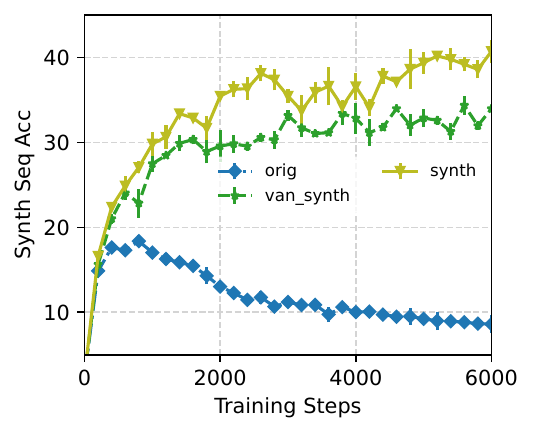}
\caption{}
\label{fig:seq_acc_synth}
\end{subfigure}
\caption{\small Sequence accuracy of error correction on the (a) original validation data  and (b) synthetic validation data. The models are trained with the original training data, synthetic training data, and vanilla sampling of synthetic data without clustering. } \label{fig:seq_acc}
\end{figure}

\begin{figure}[htb]
\centering
\begin{subfigure}[b]{0.45\linewidth}
\centering
\includegraphics[width=\textwidth]{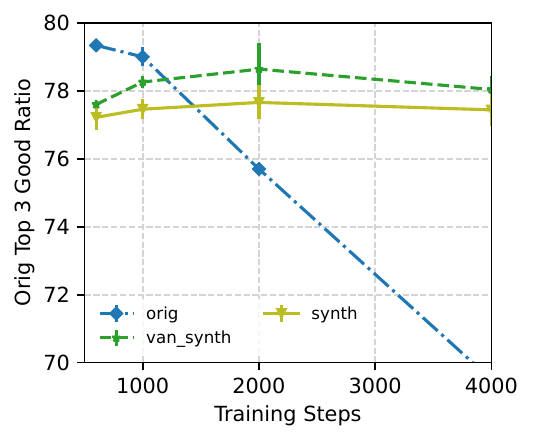}
\caption{\small }
\label{fig:top3_orig}
\end{subfigure}
\begin{subfigure}[b]{0.45\linewidth}
\centering
\includegraphics[width=\textwidth]{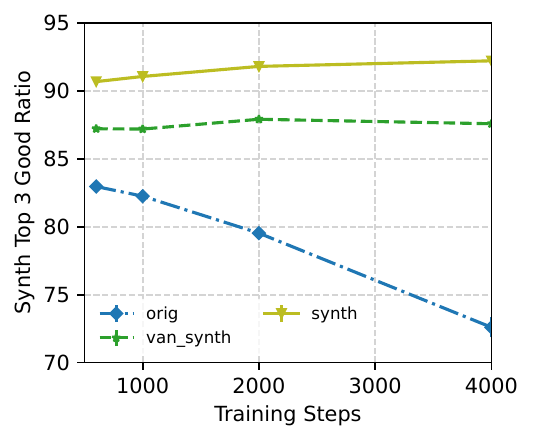}
\caption{\small }
\label{fig:top3_synth}
\end{subfigure}
\caption{Good ratio for the best of top 3 candidates for error correction on the (a) original validation data and (b) synthetic validation data. The models are trained with the original training data, synthetic training data, and vanilla sampling of synthetic data without clustering. LLMs are used to judge whether the error correction output is acceptable to compute good ratio. } \label{fig:top3}
\end{figure}
\clearpage

\section{Additional results in \cref{sec:mix}} \label{app:mix_results}

\begin{figure}[htb]
\centering
\begin{subfigure}[b]{0.45\linewidth}
\centering
\includegraphics[width=\textwidth]{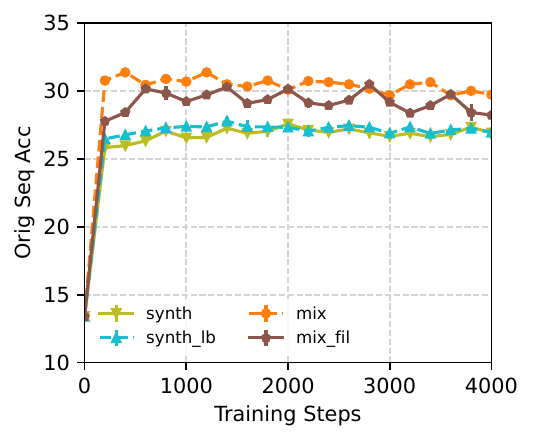}
\caption{\footnotesize}
\label{fig:seq_acc_orig_lb}
\end{subfigure}
\begin{subfigure}[b]{0.45\linewidth}
\centering
\includegraphics[width=\textwidth]{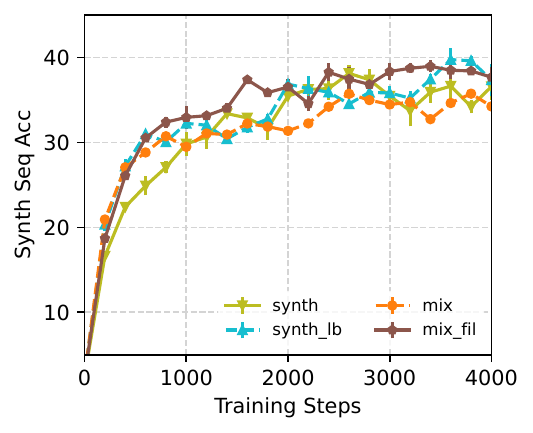}
\caption{\footnotesize}
\label{fig:seq_acc_synth_lb}
\end{subfigure}
\caption{\small Sequence accuracy of error correction on the (a) original validation data  and (b) synthetic validation data. The models are trained with synthetic training data with the same setting as in \cref{fig:top3}; $\times 4$ increased batch size (synth\_lb); mixture of original and synthetic data; and mixture of original and filtered by $w(\theta, \cdot)$ (mix\_fil). } \label{fig:seq_acc_lb}
\end{figure}

\begin{figure}[htb]
\centering
\begin{subfigure}[b]{0.45\linewidth}
\centering
\includegraphics[width=\textwidth]{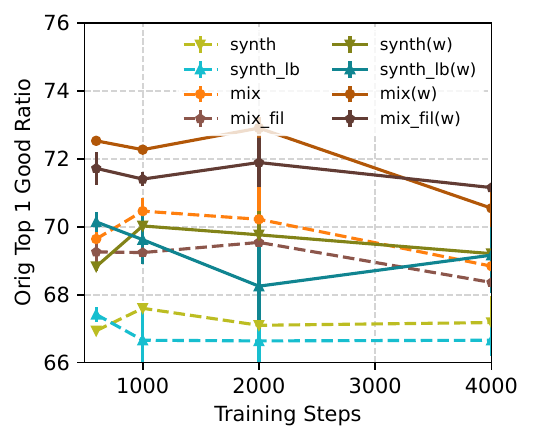}
\caption{\footnotesize }
\label{fig:top1_orig_lb}
\end{subfigure}
\begin{subfigure}[b]{0.45\linewidth}
\centering
\includegraphics[width=\textwidth]{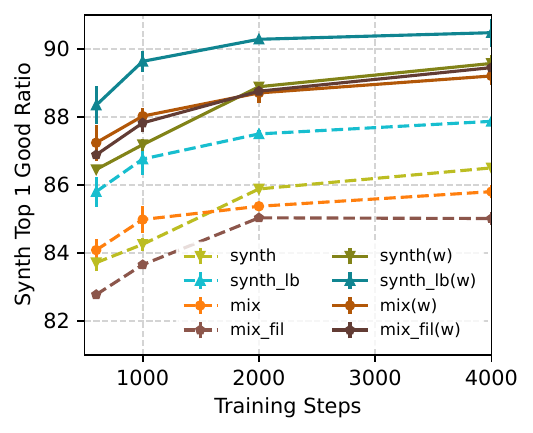}
\caption{\footnotesize }
\label{fig:top1_synth_lb}
\end{subfigure}
\caption{\small Good ratio for error correction on the (a) original validation data  and (b) synthetic validation data. Solid lines reweight the samples by the $w(\theta, \cdot)$ model learnt to fit live A/B test metrics in \cref{sec:adapt}. The models are trained with synthetic training data with the same setting as in \cref{fig:top3}; $\times 4$ increased batch size (synth\_lb); mixture of original and synthetic data; and mixture of original and filtered by $w(\theta, \cdot)$ (mix\_fil). LLMs are used to judge whether the error correction output is acceptable to compute good ratio.} \label{fig:top1_lb}
\end{figure}

\ifthenelse{\boolean{acl}}{
\section{Limitation and Future Work}
Our preliminary exploration on reweighting suggests combining live data from production applications and LLMs in a privacy-preserving manner is promising, and there are a lot of possibilities with the limited accessible information. Our usage of reweighting scores in training by filtering samples considers the trade-off of effectiveness, easy-to-implement, and future maintenance in production. There are many other potential domain adaptation methods for future experiments. Finally, We leverage a small privacy-preserving LM to capture domain information from mobile applications, while other forms of information such as histogram~\citep{xie2024differentially,hou2024pre,yu2024privacy,tan2025synthesizing} are worth considering, especially given the flexibility of the next generation FL systems in trusted execution environments~\citep{daly2024federated}. With more data generated from the interaction of LLMs and users in production deployment, our approach can become more powerful for not only domain adaptation but also other improvement such as personalization and agency, enabled by privacy-preserving methods.  
}{}

\end{document}